\documentclass[letterpaper, 10 pt, conference]{IEEEtran}
\IEEEoverridecommandlockouts
\usepackage[T1]{fontenc}
\usepackage{textcomp}
\usepackage[bookmarks=true]{hyperref}
\usepackage{amsmath}
\usepackage[capitalise]{cleveref}
\usepackage{siunitx}
\usepackage[numbers, sort]{natbib}
\usepackage{amsmath,amssymb,amsfonts}
\usepackage{algorithmic}
\usepackage{graphicx}
\usepackage{enumerate}
\usepackage{textcomp}
\usepackage{xcolor}
\usepackage{cuted}
\usepackage{subcaption}
\usepackage{glossaries}
\usepackage{booktabs}
\usepackage{pifont}
\usepackage{adjustbox} 
\usepackage{bm}
\def\BibTeX{{\rm B\kern-.05em{\sc i\kern-.025em b}\kern-.08em
    T\kern-.1667em\lower.7ex\hbox{E}\kern-.125emX}}
\usepackage{eso-pic}
\newcommand\AtPageUpperCenterNotice[1]{%
  \AtPageUpperLeft{
    \put(\LenToUnit{0.5\paperwidth},\LenToUnit{-2cm}){\makebox[0pt]{#1}}
  }
}

\AddToShipoutPictureBG*{%
  \AtPageUpperCenterNotice{%
    \parbox[b][2cm][c]{\paperwidth}{
      \centering
      \fontsize{12}{14}\selectfont
      \color{gray!50}
      This paper has been accepted for publication at the\\
      10th IEEE International Workshop on Advances in Sensors and Interfaces (IWASI) \copyright{}IEEE.
    }
  }
}

\AddToShipoutPictureBG*{
  \AtPageLowerLeft{%
    \raisebox{25pt}{\makebox[\paperwidth]{\begin{minipage}{21cm}\centering
    \fontsize{10}{8}\selectfont
    \textcolor{gray!50}{%
      \copyright{} 2025 IEEE. Personal use of this material is permitted.
      Permission from IEEE must be obtained for all other uses, in any current or future media,
      including reprinting/republishing this material for advertising or promotional purposes,
      creating new collective works, for resale or redistribution to servers or lists,
      or reuse of any copyrighted component of this work in other works.
    }
    \end{minipage}}}%
  }
}

\newcommand{\cmark}{\ding{51}}%
\newcommand{\xmark}{\ding{55}} 
\newacronym{llm}{LLM}{Large Language Model}
\newacronym{vlm}{VLM}{Vision Language Model}
\newacronym{vla}{VLA}{Vision Language Action}
\newacronym{rag}{RAG}{Retrieval Augmented Generation}
\newacronym{lora}{LoRA}{Low Rank Adaptation}
\newacronym{mpc}{MPC}{Model Predictive Controller}
\newacronym{rmse}{RMSE}{Root Mean Square Error}
\newacronym{hmi}{HMI}{Human Machine Interaction}
\newacronym{ads}{ADS}{Autonomous Driving Systems}
\newacronym{ml}{ML}{Machine Learning}
\newacronym{nn}{NN}{Neural Network}
\newacronym{rl}{RL}{Reinforcement Learning}
\newacronym{ai}{AI}{Artificial Intelligence}
\newacronym{obc}{OBC}{OnBoard Computer}
\newacronym{gpu}{GPU}{Graphics Processing Unit}
\newacronym{cpu}{CPU}{Central Processing Unit}
\newacronym{ram}{RAM}{Random Access Memory}
\newacronym{ros}{ROS}{Robot Operating System}
\newacronym{peft}{PEFT}{Parameter Efficient Fine-Tuning}
\newacronym{cot}{CoT}{Chain of Thought}
\newacronym{sft}{SFT}{Supervised Fine-Tuning}
\newacronym{gp}{GP}{Gaussian Process}
\newacronym{lidar}{LiDAR}{Light Detection and Ranging}
\newacronym{ftg}{FTG}{Follow-The-Gap}
\newacronym{fsd}{FSD}{Formula Student Driverless}
\newacronym{iac}{IAC}{Indy Autonomous Challenge}
\newacronym{sota}{SotA}{State of the Art}
\newacronym{map}{MAP}{Model- and Acceleration-based Pursuit}
\newacronym{dtr}{DTR}{Delaunay Triangulation-based Racing}
\newacronym{dwa}{DWA}{Dynamic Window Approach}
\newacronym{erpm}{ERPM}{Electric Revolutions Per Minute}
\newacronym{imu}{IMU}{Inertial Measurement Unit}
\newacronym{rrt}{RRT}{Rapidly exploring Random Tree}

\begin{document}

\title{DTR: Delaunay Triangulation-based Racing for Scaled Autonomous Racing}

\author{Luca Tognoni, Neil Reichlin, Edoardo Ghignone, Nicolas Baumann, Steven Marty, Liam Boyle, Michele Magno}
\maketitle

\begin{strip}
\vspace{-1.0cm}
\centering
\includegraphics[angle=0,origin=c,width=\textwidth]{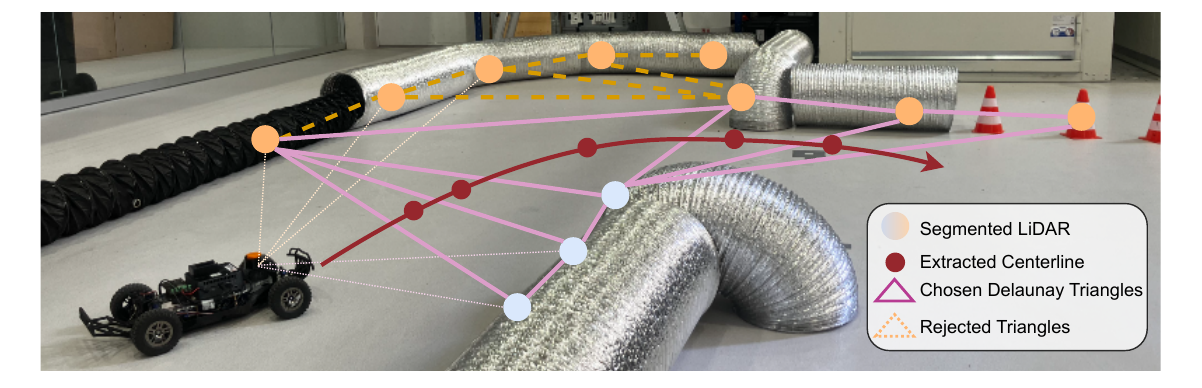}
\vspace{-0.25cm}
\captionof{figure}{Graphical illustration of the proposed \gls{dtr} reactive controller, which uses \emph{Delaunay} triangulation to extract the centerline with segmented track boundaries directly from \gls{lidar} measurements. This segmentation enables the proposed algorithm to circumvent dead-ends, a common failure case in the reactive \gls{ftg} method. Additionally, the velocity target is computed using the curvature of the extracted line and an estimated friction level  
to achieve a practical approximation of the maximum available grip.}
\label{fig:graphical_abstract}
\vspace{-0.25cm}
\end{strip}
\glsresetall

\begin{abstract}
Reactive controllers for autonomous racing avoid the computational overhead of full \emph{See-Think-Act} autonomy stacks by directly mapping sensor input to control actions, eliminating the need for localization and planning. A widely used reactive strategy is \gls{ftg}, which identifies gaps in \gls{lidar} range measurements and steers toward a chosen one. While effective on fully bounded circuits, \gls{ftg} fails in scenarios with incomplete boundaries and is prone to driving into dead-ends, known as \gls{ftg}-traps.
This work presents \gls{dtr}, a reactive controller that combines \emph{Delaunay} triangulation, from raw \gls{lidar} readings, with track boundary segmentation to extract a centerline while systematically avoiding \gls{ftg}-traps. Compared to \gls{ftg}, the proposed method achieves lap times that are 70\% faster and approaches the performance of map-dependent methods. With a latency of 8.95 ms and \gls{cpu} usage of only 38.85\% on the robot's \gls{obc}, \gls{dtr} is real-time capable and has been successfully deployed and evaluated in field experiments.
\end{abstract}


\section{Introduction}
Autonomous racing acts as a frontier for general \gls{ads}, with several autonomous racing leagues emerging from academic initiatives to advance the field \cite{catalyst0, av_survey}, such as F1TENTH \cite{okelly2020f1tenth}, \gls{iac} \cite{indyautonomous}, and \gls{fsd} \cite{amz_fullstack}. Typically, these races are conducted in structured environments, where a closed track is provided, and robots must traverse a premapped track, which enables the use of offline global mapping techniques to compute the optimal lap time a-priori, allowing precise determination of the racing line and optimal velocity profile \cite{forzaeth}. This work builds upon the open-source implementation provided by \cite{forzaeth}, utilizing a 1:10 scale autonomous race car equipped with fully commercial off-the-shelf hardware. This practical setup enables reproducible and cost-effective benchmarking of real-time autonomous driving algorithms, making it ideal for evaluating lightweight, reactive methods such as the one proposed in this work and comparing them against \gls{sota} controllers relying on full map knowledge \cite{becker2023map, forzaeth}. 

Most high-performance \gls{ads} solutions, including those used in \gls{fsd} and \gls{iac}, rely on a full \emph{See-Think-Act} autonomy stack \cite{siegwart_amr}. These systems decompose autonomy into perception, planning, and control modules, and often depend on accurate localization and global map information \cite{forzaeth}. While such pipelines enable optimal racing line tracking with full model dynamics knowledge, they are computationally intensive and require detailed prior knowledge of the environment. This limits their deployment in unknown or unstructured settings, particularly on resource-constrained \glspl{obc} deployed on mobile robots.

In contrast, reactive methods such as \cite{ftg} do not require global knowledge or full-state estimation. Instead, they infer control actions directly from local sensor measurements. Among others, \gls{ftg} \cite{ftg} is a lightweight gap-following approach enabling track traversal by identifying the largest open space in \gls{lidar} range data and steering toward it. By adjusting velocity based on steering angle and tuning associated parameters, competitive lap times can be achieved, especially through track-specific overfitting. While computationally efficient, \gls{ftg} lacks global awareness and is susceptible to suboptimal behavior, particularly in environments with dead-ends or concave boundaries, which we refer to as \gls{ftg}-traps within this work. 

This work introduces \gls{dtr} a reactive controller that operates directly on range sensor readings such as \gls{lidar} or potentially radar. It employs \emph{Delaunay} triangulation to extract the road centerline and introduces a segmentation-based approach to mitigate \gls{ftg}-traps. Furthermore, it incorporates friction- and curvature-based limits to compute an effective velocity target based on the extracted centerline. The contributions of this work are summarized as follows:

\begin{enumerate}[I] 
\item \textbf{No Dead-Ends}: Classical centerline extraction is combined with segmentation and track geometry knowledge to guide the \emph{Delaunay} triangulation, preventing traversal into dead-ends and oscillating behaviour, compared to classical \gls{ftg}.

\item \textbf{Comparison}: The proposed algorithm is evaluated against classical \gls{ftg} and other \gls{sota} controllers that assume full map knowledge. This enables comparisons to current reactive systems and controllers with access to global information. 
Being only 38\% slower (in terms of lap-time) than \gls{sota} algorithms, our approach sets a new standard for reactive racing and outperforms \gls{ftg}, which is instead 126\% slower than the baseline. 

\item \textbf{Real World Deployment}: The method is deployed in a closed control loop directly on the robot’s \gls{obc}. A latency of 8.95 ms average is achieved, demonstrating computational feasibility even on resource-constrained robotic \glspl{obc} such as the Intel Core i5-10210U \gls{cpu}. 
\end{enumerate}

\section{Related Work}
High-speed autonomous racing has been extensively studied as a benchmark for advanced motion planning and control. The most competitive systems typically rely on \textbf{model-based methods} that assume global map availability and precise localization, enabling offline trajectory optimization and predictive control.
However, in mapless settings---where no prior knowledge of the track is available---systems must operate solely on onboard sensing and local information.
Approaches to this challenge broadly fall into two categories: \textbf{end-to-end learning} methods that infer control policies directly from raw sensory input, and \textbf{reactive planning} methods that rely on geometric reasoning from range measurements.

\textbf{Model-Based Methods and Trajectory Optimization}:
Among the most established and high-performing approaches to autonomous racing are model-based and optimization-driven methods.
\gls{mpc} techniques have been widely used in autonomous racing scenarios. 
\citet{williams2017information} proposed an information-theoretic \gls{mpc} framework for racing under uncertainty, while \citet{liniger2015optimizationbased} presented an optimization-based controller capable of performing aggressive maneuvers with high-fidelity vehicle models.  \citet{hewing2020cautious} further incorporated \gls{gp} models into nonlinear \gls{mpc} to improve robustness in the presence of model uncertainty.
In the context of F1TENTH, \citet{becker2023map} implemented \gls{map}, a model-based controller that achieves \gls{sota} performance by tracking a precomputed racing line. 

These methods typically rely on a global track map, an offline-optimized reference trajectory, and accurate localization through dense state estimation—assumptions that do not hold in fully mapless settings. Furthermore, their dependence on complex online optimization and full system modeling imposes high computational demands, making them less suitable for real-time racing on resource-constrained platforms and placing them in a fundamentally different paradigm than our approach. Instead of planning over a global track model, we generate trajectories directly from raw \gls{lidar} geometry, avoiding the need for an autonomy stack relying on full state estimation and enabling a simpler system that still supports structured behavior in real time.

\textbf{End-to-End Learning:} 
As an alternative to model-based approaches, end-to-end learning methods directly map sensor inputs to control actions, enabling mapless autonomous driving.
\citet{kendall2019learning} demonstrated a deep \gls{rl} system capable of driving using only monocular images, showing initial success but requiring substantial data collection and training time. 
A comprehensive survey by \citet{betz2022survey} summarized end-to-end autonomous racing approaches, noting that many have shown promising results, particularly in simulated environments, and \cite{zhang2022residual} specifically shows that mapless \gls{rl} algorithms have good potential in the F1TENTH simulator, albeit with no sim-to-real analysis.
Regarding high-performance racing in the physical world, the domain of autonomous quadrotor flight offers different solutions: \citet{Kaufmann2020deepdrone} performed high-speed acrobatic maneuvers and \cite{loquercio2021learning} achieved collision-free navigation in complex natural and human-made environments---both using end-to-end \gls{rl} and leveraging privileged simulation information to bridge the sim-to-real gap.
While these methods show remarkable adaptability, they offer limited transparency, require careful tuning, and typically do not exploit the structured geometry of race tracks. 
In contrast, our approach combines the interpretability and efficiency of classical planning with structured reasoning over the traversable environment, enabling safe and aggressive behavior without relying on learned models.

\textbf{Reactive Planning}:
In contrast to learning-based approaches, classical reactive methods offer lightweight, geometry-driven solutions for mapless navigation by responding directly to local sensor observations. The F1TENTH platform \cite{okelly2020f1tenth} popularized real-time gap-finding based purely on 2D \gls{lidar} scans, enabling basic autonomous racing without a global map, commonly known as the \gls{ftg} method \cite{ftg}. The \gls{dwa} \cite{fox1997dwa} similarly computes feasible short-horizon velocity commands to avoid collisions while progressing toward a goal.
\citet{feraco2020local} proposes a planner based on \gls{rrt} \cite{karaman2010rrt}, which however was only tested in a simulated environment.

\textbf{Our Contribution}:
We propose a geometric trajectory generation approach for mapless autonomous racing, leveraging real-time 2D \gls{lidar} scans.
\emph{Delaunay} triangulation is used to create a structured representation of the scanned environment, explicitly modeling track boundary segments. 
A dynamic velocity target is generated, according to the assumed friction level and the target curvature, enabling aggressive yet safe high-speed navigation.
Our approach requires no prior map, minimal tuning, and runs fully in real time.
Our technique offers greater interpretability, predictable behavior, and higher data efficiency than end-to-end learning methods. Compared to gap-finding reactive methods, it achieves smoother and faster racing performance by better exploiting the local geometrical structure.
Through extensive experimental validation, we demonstrate state-of-the-art performance in fully mapless autonomous racing, achieving speeds and robustness levels exceeding previous mapless approaches. A table comparing the proposed method against related work is available in \Cref{tab:rw}.

\begin{table} [h]
\centering
    \begin{adjustbox}{max width=\columnwidth}
    \begin{tabular}{l|c|c|c|c}
        \toprule
          & \textbf{Mapless} & \textbf{Sensor Setup} & \textbf{Real Time} & \textbf{Speed Deficit}\\
        \midrule
        \acrshort{rrt}-based \cite{feraco2020local} & \cmark & 3D \gls{lidar} & \xmark & N/A \\
        \acrshort{ftg} \cite{ftg, forzaeth}  & \cmark & 2D \gls{lidar} & \cmark & 126.1\% \\
        \acrshort{dtr} \textbf{(Ours)}  & \cmark & 2D \gls{lidar} & \cmark & 38.2\% \\
        \midrule
        \acrshort{map} \cite{becker2023map}  & \xmark & \textit{Full-Stack} & \cmark & - \\
        \acrshort{mpc} \cite{williams2017information, liniger2015optimizationbased, hewing2020cautious} & \xmark & \textit{Full-Stack} & \cmark & - \\
        \bottomrule
    \end{tabular}
    \end{adjustbox}
\caption{Comparison of representative racing algorithms. Methods are grouped by whether they rely on a global map. \textbf{Mapless} indicates methods that operate without prior map information. \textbf{Sensor Setup} specifies the main onboard sensor. \textbf{Real Time} denotes whether the method was deployed and evaluated on a physical platform. \textbf{Speed Deficit} quantifies the relative increase in lap time, expressed as a percentage, compared to the best-performing (fastest and map-based) controller \gls{map} \cite{becker2023map} tested on the F1TENTH platform. Note that this comparison is limited to mapless methods; for map-based approaches, lap times are not reported in this metric as they serve as the reference baseline.}
\label{tab:rw}
\end{table}


\section{Methodology}
This work builds upon the open-source implementation provided by \cite{forzaeth}, utilizing a 1:10 scale autonomous race car equipped with fully commercial off-the-shelf hardware. The primary onboard sensors include a \gls{imu}, wheel speed sensing via \gls{erpm}, and a \gls{lidar} sensor, which serves as the main exteroceptive input by providing range measurements critical to this approach. The \gls{ftg} baseline implementation from \cite{forzaeth} is directly adopted and serves as a reference throughout this work.

\begin{figure*}[!htb]
    \centering
    \begin{subfigure}[b]{0.49\textwidth}
        \centering
        \includegraphics[height=5cm]{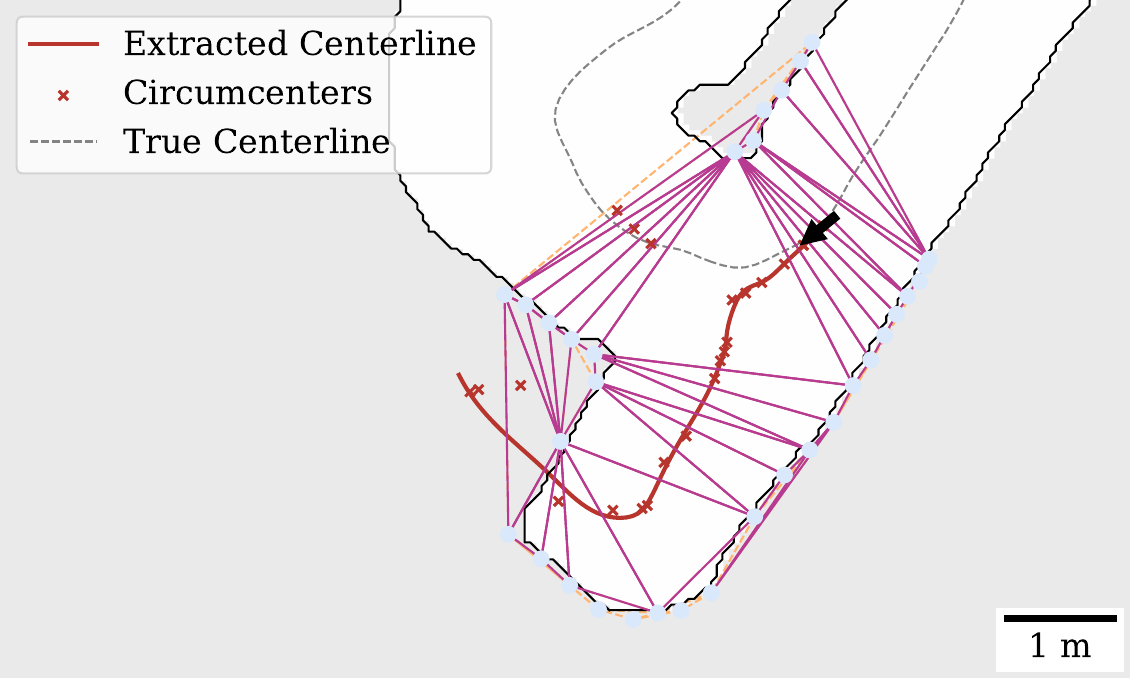}
        \caption{Raw Centerline Extraction}
        \label{fig:centerline}
    \end{subfigure}
    \hfill
    \begin{subfigure}[b]{0.49\textwidth}
        \centering
        \includegraphics[height=5cm]{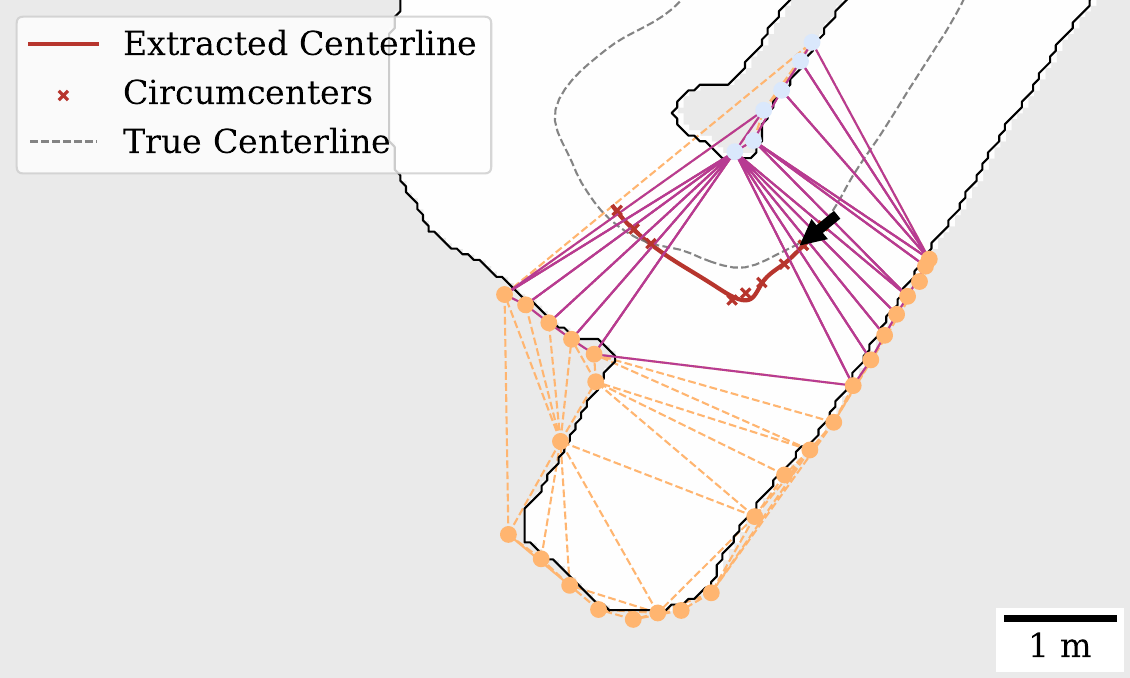}
        \caption{Centerline Extraction with Wall Segmentation}
        \label{fig:wall_segment}
    \end{subfigure}
    \caption{Visualization of the \emph{Delaunay} triangulation of the same setup as in \cref{fig:graphical_abstract}. The current position of the race car is indicated by a black arrow. Triangles that were filtered out by the geometric heuristics are shown with dashed edges. (a) Centerline extraction through triangulation of raw range measurements (b) Wall segmentation to ensure only triangles with vertices containing at least two classes of segments are used (e.g., inner track segment is light-blue, outer track segment in orange). This way \gls{ftg}-traps such as dead-ends can be systematically rejected.}
    \label{fig:triangulation}
\end{figure*}

\subsection{Centerline Extraction}
Similarly to the approach of \citet{amz_fullstack}, we use \emph{Delaunay} triangulation to extract the centerline from range measurements of the track boundaries. For that, the 2D \gls{lidar} scans are subsampled according to the \emph{boxed} approach in \cite{stahl2019ros}. The \emph{Delaunay} Triangulation is then performed on these \gls{lidar} points, connecting them to triangles such that none of the points lie within the circumcircle of the triangles.
To ensure that only meaningful triangles contribute to the centerline extraction, we apply several geometric heuristics to the \emph{Delaunay} triangulation of the \gls{lidar} scan points:

\begin{itemize}
  \item \textbf{Isosceles-like Condition:} A triangle is considered \emph{isosceles-like} if its two longer sides are of similar length.
  
  \item \textbf{Pointedness Condition:} A triangle is considered \emph{pointed} if the ratio between its longest and shortest side is sufficiently large.
  
  \item \textbf{Area Condition:} A triangle is considered \emph{large} if its area exceeds a threshold.
\end{itemize}

A triangle is retained if it satisfies either both the \emph{isosceles-like} and \emph{pointedness} conditions, or the \emph{area} condition.
These heuristics ensure that only triangles likely to span the track width are used for centerline extraction.

With the set of circumcenters identified, the next step is to connect them into an approximate centerline, which can later serve as the racing line. To ensure that only relevant points are used for this, we apply additional filtering, retaining only those circumcenters that lie ahead of the race car and within the \gls{lidar} scans.
The centerline is then constructed using a greedy nearest-neighbor algorithm. Starting at the circumcenter closest to the car's current position, the algorithm iteratively selects the nearest unvisited circumcenter that satisfies two constraints: it must not lie significantly behind the current point in the driving direction, and it must not exceed a predefined maximum distance. This process continues until no further valid points remain. The result is an ordered sequence of waypoints that forms a preliminary estimate of the track’s centerline.

The ordered centerline points are smoothed using a Savitzky-Golay filter. A spline is then fitted to the smoothed points, resulting in a continuous and smooth approximation of the centerline, which will be used as the racing line. Additionally, the curvature of the resulting spline is computed and later used for the velocity computation. The centerline extraction result shown in \Cref{fig:centerline} demonstrates how the method can mistakenly lead the car into a dead-end. In this particular case, although the generated path initially follows the center of the corridor, the algorithm continues straight into the dead-end. This happens because a cluster of circumcenters exists in that direction, and the greedy nearest-neighbor search connects to them, unaware that the path cannot continue. As a result, the extracted racing line goes into the dead-end rather than following the intended racetrack.

\subsection{Wall Segmentation}
By segmenting the \gls{lidar} range measurements, illustrated in \Cref{fig:wall_segment}, into consecutive track boundaries using distance thresholding, the subsequent \emph{Delaunay} triangulation is restricted to use only the circumcenters of triangles whose vertices span at least two distinct boundary segment classes. This constraint ensures that dead-ends are avoided, as such regions would produce triangles with vertices belonging solely to a single class. Consequently, the raw centerline extraction shown in \Cref{fig:centerline} is improved, enabling systematic avoidance of \gls{ftg}-traps.

\subsection{Velocity Profile Computation}
Once a centerline is available, the controller computes the control commands—namely, steering angle and target velocity—that guide the race car along the track. To do so, it first determines a suitable lookahead point on the centerline, located a short distance ahead of the vehicle. This lookahead distance scales linearly with the current speed, ensuring that the controller plans further ahead at higher speeds, which enables smoother and more stable tracking.
Based on the selected lookahead point along the centerline, the steering angle is computed using the kinematic bicycle model. The model assumes a single-track approximation of the racecar, and the steering command is derived geometrically by computing the angular deviation between the vehicle’s current heading and the direction of the lookahead point. This angle is then transformed into a steering angle using the bicycle model's formulation, similarly to \cite{pure_pursuit}.

The velocity commands are computed based on the vehicle's dynamic constraints, ensuring safe navigation through curves by accounting for track curvature and available tire-road grip. 
First, an estimate of the friction coefficient $\mu$ is obtained by pulling the car laterally along the center of mass with a spring scale.
The maximum admissible speed $v_{adm}$ is then derived from the balance between lateral acceleration --- directly influenced by the local curvature --- and the frictional limits of the vehicle, inspired by the first approximation used for velocity generation in \cite{heilmeier2020mincurv}:
\begin{equation*}
    v_{adm} = \sqrt{\mu\,a_y^{max}\,\kappa^{-1}}
\end{equation*}
where $a_y^{max}$ is an underestimation of the maximum lateral acceleration and $\kappa$ is the unsigned curvature. 
This formulation ensures that the commanded speed remains within the bounds of what the tires can physically sustain without loss of traction. In practice, the race car reduces its speed in regions of high curvature and is permitted to accelerate in flatter segments.
Speed transitions are further constrained by predefined longitudinal acceleration and deceleration limits, yielding smooth and dynamically feasible velocity profiles along the predicted path.

\section{Results}
\subsection{Comparison of Lap Time and Computational Performance}
\Cref{tab:laptimes} compares our reactive controller (\gls{dtr}) with the classical \gls{ftg} and two non-reactive, map-based controllers (\gls{map} and \gls{mpc}). \gls{dtr} significantly improves over \gls{ftg}, reducing lap time from 9.45s to 5.79s, showing the effectiveness of our centerline-based approach. 
Compared to \gls{map} and \gls{mpc}, which rely on global racing lines and full track knowledge, \gls{dtr} is still significantly slower. This comparison highlights a central trade-off: while reactive methods typically sacrifice performance for autonomy, \gls{dtr} closes much of the performance gap to map-based controllers without needing any a priori information.
 
\begin{table} [h]
\centering
\begin{adjustbox}{max width=\columnwidth}
\begin{tabular}{l|c|c||c|c}
    \toprule
     & \textbf{\acrshort{dtr} (Ours)} & \textbf{\acrshort{ftg}} & \textbf{\acrshort{map}} & \textbf{\acrshort{mpc}} \\ \midrule
    \bm{$t_{lap}$} [s] $\downarrow$ & $5.79 \pm 0.11$ & $9.45 \pm 0.31$ & $4.19 \pm 0.03$ & $4.18 \pm 0.01$ \\ 
    \bm{$\tau$} [ms] $\downarrow$ & $8.95 \pm 2.34$ & $1.18 \pm 0.22$ & $1.38 \pm 0.23$ &$ 16.86 \pm 3.88$ \\ 
    \bm{$U$} [\%] $\downarrow$ & $38.86 \pm 2.04$ & $9.29 \pm 0.62$ & $12.35 \pm 1.33$ & $68.27 \pm 4.75$ \\ 
    \textit{Reactive}  & \cmark & \cmark & \xmark & \xmark \\ 
    \bottomrule
\end{tabular}
\end{adjustbox}
\caption{Performance comparison between \textit{Reactive} controllers \gls{dtr} (ours) and \gls{ftg}, and non-reactive \gls{sota} controllers \gls{map} and \gls{mpc}, which necessitate full track knowledge to track the global racing line and incorporate model dynamics.
All metrics are reported in the $\mu \pm \sigma$ format  ($\mu$: mean, $\sigma$: standard deviation). 
Reported lap times $t_{lap}$ are accumulated over 5 laps (only laps without collisions are considered). All controllers were manually tuned to achieve the lowest possible lap time. 
The latency and \gls{cpu} utilization of the different controllers are denoted with $\tau$ and $U$, respectively. Given our setup, the maximum amount for $U$ is $800\%$.}
\label{tab:laptimes}
\end{table}

\begin{figure*}[htb] 
    \centering
    \includegraphics[width=\linewidth, trim=0cm 0cm 0cm 0cm, clip]{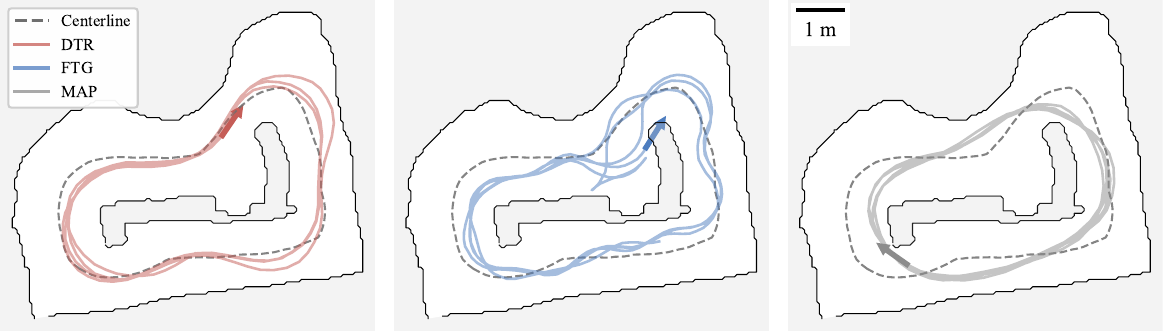}
    \caption{Qualitative comparison of the trajectories from left to right, driven by \gls{dtr} (red), \gls{ftg} (blue), and the non-reactive \gls{map} (gray) tracking the global racing line. The gray dotted line represents the centerline, and the racing line used by \gls{map} optimizes for the minimum curvature.}
    \label{fig:quali_trajectories}
\end{figure*}

Computational results highlight that this performance increase comes at a cost: namely, \gls{dtr} has a 7.6-times higher latency and a 4.2-times higher \gls{cpu} utilization compared to \gls{ftg}.
However, the overall computational requirements remain lower than the computationally intensive \gls{mpc} controller, and further \gls{dtr} does not require a state estimator module. It is indeed relevant that the reported \gls{cpu} utilization and latency reported for \gls{mpc} are only relative to the controller part: as reported from \cite{forzaeth}, the state estimation module would add a further 180\% \gls{cpu} utilization, bringing the overall approximate savings of our approach to 210\% of \gls{cpu} utilization, when compared to an \gls{mpc} approach.

\subsection{Qualitative Assessment with Trap}

As shown in \Cref{fig:quali_trajectories}, the trajectories of the reactive controllers \gls{dtr} (red) and \gls{ftg} (blue) are compared to the non-reactive \gls{map} controller (gray), which utilizes full map knowledge and follows a minimum curvature racing line. The centerline is indicated as a gray dashed line. These trajectories correspond to the quantitative lap time results presented in \Cref{tab:laptimes}.

Compared to \gls{ftg}, the proposed \gls{dtr} demonstrates significantly improved stability, producing consistent trajectories across laps. Most critically, it avoids the \gls{ftg}-trap located in the middle of the track, a result of the enhanced centerline extraction based on \emph{Delaunay} triangulation. Furthermore, \gls{dtr} closely tracks the extracted centerline, whereas \gls{ftg} exhibits notable lateral oscillations.

In comparison to \gls{map}, a \gls{sota} controller with full map access and a precomputed minimum curvature racing line, \gls{dtr} is understandably slower and less precise, as also reflected in \Cref{tab:laptimes}. However, \gls{map} requires a full \emph{See-Think-Act} architecture, including precise localization, which is often the most computationally demanding component of autonomous driving systems \cite{forzaeth}. In contrast, \gls{dtr}, being fully reactive, achieves competitive performance with substantially lower system complexity.

\section{Conclusion}

In this study, we introduced \acrfull{dtr}, a reactive end-to-end racing algorithm that improves upon the classical \gls{ftg} approach by enabling faster, more consistent driving while effectively avoiding dead-ends. With a lap time of $5.79$ seconds — significantly faster than the $9.45$ seconds achieved by \gls{ftg} — \gls{dtr} demonstrates substantial performance gains, narrowing the gap to map-based controllers despite operating without a precomputed racing line. The proposed approach introduces higher computational demands, with 7.6 times the latency and 4.2 times the \gls{cpu} usage compared to the classical \gls{ftg}. Nevertheless, it remains significantly more lightweight than \gls{mpc} methods and operates without the need for a dedicated state estimation module. Its moderate resource footprint makes it suitable for onboard deployment, including on embedded systems with limited processing capabilities.

Currently, one limitation of the approach is its limited obstacle avoidance capabilities, and future work could integrate ad-hoc filters to appropriately avoid static obstacles and dynamic opponents. Furthermore, higher-performance model-based controllers could be integrated in \gls{dtr}, such as \gls{map} or \gls{mpc} \cite{liniger2015optimizationbased}. However, an accurate model requires system identification procedures, which either require large unhindered spaces or a fully mapped environment \cite{dikici2025ontrack}.

\bibliographystyle{IEEEtranN}
\bibliography{ref}

\end{document}